\documentclass{article}

\usepackage[nonatbib,final]{neurips_2020}
\usepackage[numbers]{natbib}

\usepackage[utf8]{inputenc} 
\usepackage[T1]{fontenc}    
\usepackage{hyperref}       
\usepackage{url}            
\usepackage{booktabs}       
\usepackage{amsfonts}       
\usepackage{nicefrac}       
\usepackage{microtype}      
\usepackage{xcolor}
\usepackage{xspace}
\usepackage{tabularx}
\usepackage{amsmath,bm,multicol,multirow}
\usepackage{amssymb}
\usepackage{cleveref}
\usepackage[pdftex]{graphicx}
\usepackage{array, makecell}
\usepackage[scientific-notation=true]{siunitx}
\usepackage{subcaption}
\usepackage{chngpage}
\usepackage{wrapfig,lipsum,booktabs}
\usepackage{pifont}
\usepackage{soul}
\usepackage{tikz}

\newcolumntype{L}{>{\raggedright\arraybackslash}X}

\usepackage{ulem}
\title{\LARGE Meta learning to classify intent and slot labels with noisy few shot examples}

\author{%
    Shang-Wen Li, Jason Krone, Shuyan Dong, Yi Zhang, Yaser Al-onaizan \\
    Amazon AI\\
    {\tt \{shangwel, kronej, shuyand, yizhngn, onaizan\}@amazon.com} \\
}

\begin{document}

\maketitle

\begin{abstract}
   Recently deep learning has dominated many machine learning areas, including spoken language understanding (SLU). However, deep learning models are notorious for being data-hungry, and the heavily optimized models are usually sensitive to the quality of the training examples provided and the consistency between training and inference conditions. To improve the performance of SLU models on tasks with noisy and low training resources, we propose a new SLU benchmarking task: few-shot robust SLU, where SLU comprises two core problems, intent classification (IC) and slot labeling (SL). We establish the task by defining few-shot splits on three public IC/SL datasets, ATIS, SNIPS, and TOP, and adding two types of natural noises (adaptation example missing/replacing and modality mismatch) to the splits. We further propose a novel noise-robust few-shot SLU model based on prototypical networks. We show the model consistently outperforms the conventional fine-tuning baseline and another popular meta-learning method, Model-Agnostic Meta-Learning (MAML), in terms of achieving better IC accuracy and SL F1, and yielding smaller performance variation when noises are present.

\end{abstract}

\section{Introduction and Related Works}

Goal-oriented dialogue systems is a hot topic in machine learning research. The systems have widespread applications in the industry and are the foundation of many successful products, including Alexa, Siri, Google Assistant, and Cortana. One core component of a dialog system is spoken language understanding (SLU), which consists of two main problems, intent classification (IC) and slot labeling (SL) \cite{old_SLU_tur, Delik_word_confusion}. In IC, we attempt to classify the goal of a user query, usually input in text or transcribed by automatic speech recognition (ASR) system from audio. SL, similar to the named-entity recognition (NER) problem, aims to label each token in a query an entity type. The only difference is that entity types in SL are domain-specific and based upon dialog ontology. Recent advances in neural models have enabled greatly improved SLU \cite{yao2014spoken, guo2014joint, mesnil2014using, cao2020style, lai2020towards}.

However, two significant challenges hinder the broad application and expansion of the SLU models in industrial settings. First of all, neural methods require a large amount of labeled data for training \cite{liu2020tera}. SLU is often coupled with the ontology of the underlying dialog system and thus domain-dependent. Collecting a large number of in-domain labeled data for neural models is prohibitively expensive and time-consuming. Secondly, the performance of SLU models in practice often suffers from fluctuations due to various types of noises. One common noise is adaptation data perturbation. In many industrial applications such as cloud services\footnote{Alexa ASK: https://developer.amazon.com/en-US/alexa/alexa-skills-kit; Google DialogFlow: https://dialogflow.com/}, the SLU model is built by fine-tuning (or adapting) a pre-trained, shared network to the target domain with data provided by developers. The developers often have a limited background in SLU and machine learning. Thus the data provided varies in quality and is subject to different types of perturbations, such as missing or replaced data samples (e.g., a subset of an optimal example set is missing or replaced by redundant examples in the adaptation data) and typos. Another common noise comes from the mismatch of input modalities between adaptation and inference stages. For instance, the model is adapted with human transcription yet deployed to understand ASR decoded text, or the input at adaptation and inference stages relies on the recognition of different versions of ASR models. Given that most neural methods comprise a large number of parameters and are heavily optimized for the training (i.e., adaptation in the context of cloud service) data provided, the resulting model is usually sensitive to these noises. The requirement of noise-free adaptation and inference conditions also prohibits the use of neural SLU techniques because it is often infeasible to achieve such conditions.

Transfer learning and meta-learning are two conventional techniques that have been applied to address the challenge of data scarcity. Transfer learning usually refers to pre-training initial models using mismatched domains with rich human annotations and then adapting the models with limited labels in targeted domains. Previous works \cite{yang2017transfer, kumar2017just, schuster2019cross, chen2019bert, lai2020semi} have shown promising results in applying transfer learning to SLU. Note that pre-training discussed here covers methods including using a pre-trained language model like BERT \cite{BERT} directly and further training downstream tasks on data in mismatched domains with the pre-trained model. In the following, we focus on the latter due to utilizing data from other domains better and yielding higher accuracy. In recent years, meta-learning has gained growing interest among the machine learning fields for tackling few-shot learning (i.e., data scarcity) scenarios. Model-Agnostic Meta-Learning (MAML) \cite{MAML} focuses on learning parameter initialization from multiple subtasks, such that the initialization can be fine-tuned with few labels and yield good performance in targeted tasks. Metric-based meta-learning, including prototypical networks (ProtoNets) \cite{Proto, Luo2020} and matching networks \cite{Match}, aim to learn embedding or metric space which can be generalized to domains unseen in the training set after adaptation with a small number of examples from the unseen domains. Recent work unveils excellent potential in applying meta-learning techniques to SLU in the few-shot learning context \cite{Jason_Yi_ACL}.

As compared to data scarcity, another challenge for SLU, the robustness against noises, is also gaining attention. Simulated ASR errors are used to augment training data for SLU models \cite{Simonnet}. Researchers also leverage information from confusion networks or lattices \cite{Delik_word_confusion, Tur_semantic_parsing, LatticeRnn, Vivian_robust_SLU, masumura2018neural, yaman2008integrative}, and adversarial training techniques \cite{adversarial_training, hy_lee} for models to learn query embeddings that are robust against ASR errors. For text input, methods have also been explored on model robustness against noises from misspelling and acronym \cite{robust_ner}. In contrast to these noise types that have gained attention, to our best knowledge, there is no prior work investigating the impact of missing or replaced examples in adaptation data. Moreover, the intersection of data scarcity and noise robustness is unexplored. Since the scarcity of labeled data and data noisiness usually co-occur in SLU applications (both reflect the difficulty of acquiring annotated data), the lack of studies in the intersectional areas hinders the use of neural SLU models and its expansion to broader use cases.

Given the deficiency, we establish a novel few-shot noisy SLU task by introducing two common types of natural noise, adaptation example missing/replacing and modality mismatch, to the previously defined few-shot IC/SL splits \cite{Jason_Yi_ACL}. The task is built upon three public datasets, ATIS \cite{ATIS}, SNIPS \cite{SNIPS}, and TOP \cite{TOP}. We further propose a noise-robust few-shot SLU model based on ProtoNets for the established task. In summary, our primary contributions are 3-fold: 1) formulating the first few-shot noisy SLU task and evaluation framework, 2) proposing the first working solution for the few-shot noisy SLU with the existing ProtoNet algorithm, and 3) in the context of noisy and scarce learning examples, comparing the performance of the proposed method with conventional techniques, including MAML and fine-tuning based adaptation.

\section{Approaches}

In this section, we dive deep into the formulation of few-shot noisy SLU tasks. We also elaborate on the method we propose to overcome this challenging task.

\subsection{Problem formulation}

\subsubsection{Few-shot SLU}

The goal of the few-shot SLU is to adapt an IC/SL classifier $f_{\theta}$ pre-trained on data, $D^{pretrain}$, in mismatch domains with rich annotation to new domains using data, $D^{test}$, which comprises few labeled examples per class. In this setting, pre-training and test splits are two disjoint class sets $L^{pretrain}$ and $L^{test}$ from $D^{pretrain}$ and $D^{test}$ respectively. Classes in $L^{pretrain}$ are used for model pre-training while those in $L^{test}$ are held out for test time to adapt the pre-trained model and evaluate generalizability. The test is done episodically, where each episode is a mini adaptation task containing a support set $S$ and a query set $Q$. Adaptation of $f_{\theta}$ to new domains is achieved with a few labeled examples provided by the support set, whereas model performance is evaluated by averaging metrics measured in each episode's query set. In our implementation, we also pre-train meta-learning methods on episodes, since a previous study showed that pre-training with matched condition yields better performance \cite{Jason_Yi_ACL}.

We follow the setup in \cite{Jason_Yi_ACL} to build support and query sets. We define $S^m = \cup_{l \in L^m} S_l$ , where $S_l = \{(x_l^i, l, t_l^i), i \in (1, ... k_s)\}$ and $m$ is either the pre-train or test split. Here $x_l^i$ is the input utterance of the $i$-th example for intent (i.e., class) $l$, $t_l^i$ is the slot label of $x_l^i$, and $k_s$ is the number of examples per class in the support set. Similarly $Q^m$ is defined as $\cup_{l \in L^m} Q_l$, where $Q_l = \{(x_l^j, l, t_l^j), j \in (1, ... k_q)\}$ and $Q_l \cap S_l = \varnothing$. In this way, we construct an episode by sampling $k_s$ and $k_q$ examples per intent $l$ in $L^{pretrain}$ and $L^{test}$ from $D^{pretrain}$ and $D^{test}$ respectively. Note that in practice $D^{test}$ consists of more than a few examples per class, and thus we downsample $D^{test}$ to resemble a few-shot learning task in each episode. Besides, for fair evaluation, we map slot labels in the query set but not in the support one to \textbf{Other}, which is excluded when we evaluate SL performance. The mapping also guarantees not updating gradient for slot labels unseen in support set during pre-training episodically.

\subsubsection{Few-shot noisy SLU}

Noises usually co-occur with data scarcity in the industrial settings of SLU. Thus we further formulate a few-shot noisy SLU task for benchmark and model development. The task is built upon the few-shot SLU described above, and the goal is to adapt the IC/SL classifier with few examples such that the resulting model can perform well and robust in new domains when noise exists. Specifically, we investigate two types of noise, adaptation example missing/replacing and modality mismatch\footnote{We select the two types of noises as they are common in cloud services, where the input modality at deployment can be different from development; the provided adaptation data and its quality can fluctuate due to developers' limited background or deletion per user privacy concerns. We plan to explore other prevalent noises, including typos and acronyms, in future work.}. 

For adaptation example missing/replacing, $S^{pretrain}$, $Q^{pretrain}$, and $Q^{test}$ are kept identical to the ones in few-shot SLU, while the support set in test split, $\overline{S^{test}}$, is the perturbed $S^{test}$. In the following experiment, we perturb $S^{test}$ by either removing $c$ examples per intent from $S^{test}$, or replacing $c$ examples per intent with ones sampled from $D^{test}$ but excluding $S^{test}$ and $Q^{test}$. We choose to study these two types of perturbation, for they are the basic operation on the example set at the utterance level. A more complicated variation can be built by combining the two. To quantify the model robustness against the missing and replacing operation, in each episode at the test stage, we adapt the pre-trained classifier with $S^{test}$ and $\overline{S^{test}}$ separately, and evaluate on $Q^{test}$ to measure the performance difference between models adapted with the original and perturbed support set. With this setup, we estimate how well and robust classifiers can perform with a network pre-trained on mismatched but rich-annotated domains as well as a small and perturbed adaptation set. Good and robust performance in such a setting is especially useful in the context of cloud services.

In modality mismatch, the goal is to benchmark performance impact when the preprocessing pipeline of data used in pre-training and adaptation is different from the one in inference. We simulate this noise type by replacing examples in $Q^{test}$ with ASR hypotheses and IC/SL annotation on them. Human transcription is still used in pre-training split and $S^{test}$ for adaptation. Since most SLU benchmarking datasets only provide IC/SL annotation on human transcription, further data processing is required. Here we adopt a common technique, noise-corrupted synthesized speech \cite{Gunter, Me}, to obtain ASR hypotheses and corresponding annotation. We first apply TTS on text input and feed the synthesized audio into ASR to generate the ASR hypothesis. Levenshtein alignment between tokens in the hypothesis and original text is then adopted to project SL annotation onto the hypothesis (note that intent labeling is not affected by ASR results). Projected SL labels are reviewed and corrected by human annotators for the experiment. In such a manner, we generate examples to measure model performance and robustness when the data preprocess pipeline, i.e., text input from human vs. audio input recognized by ASR, is mismatched.

\subsection{Learning frameworks}

\begin{figure*}[t]
  \centering
  \includegraphics[width=\linewidth]{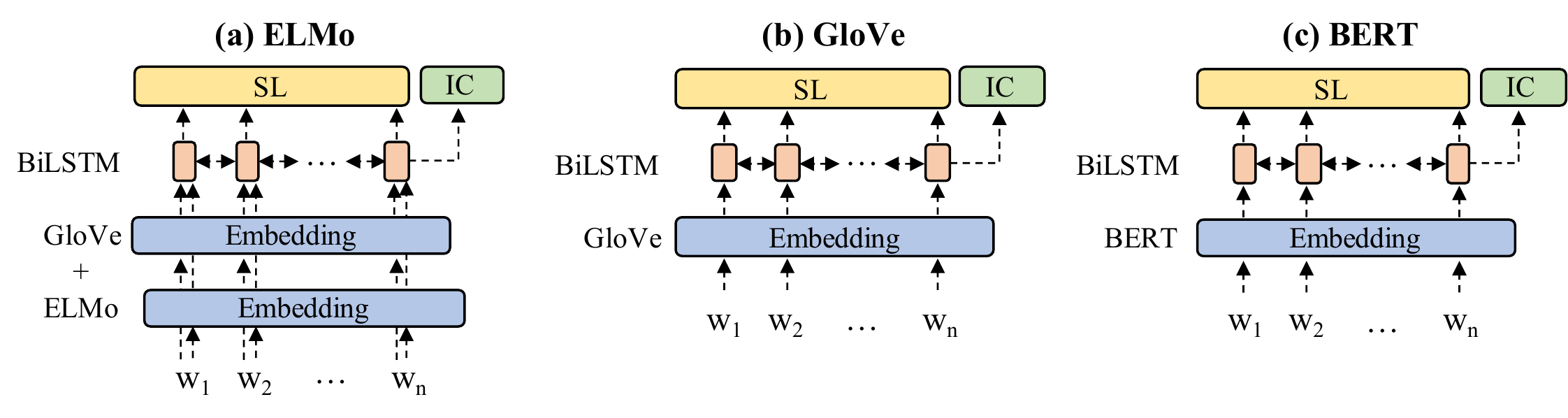}
  \caption{The architectures for the IC-SL joint classifiers built with various learning frameworks. Three popular architectures for SLU classifiers, \textbf{ELMo}, \textbf{GloVe}, and \textbf{BERT}, are shown.}
  \label{fig:diagram}
\end{figure*}

To build robust models for the few-shot noisy SLU, we propose ProtoNets \cite{Proto} based SLU. ProtoNets is a popular meta-learning framework for the few-shot learning scenario. We apply ProtoNets to IC and SL problems by first representing the input utterance and tokens with encoder $Enc(x) :\to \mathbb{R}^D$. Then we compute prototypes, i.e., centroids of examples, of each intent and slot class with the embedding. That is 
\begin{equation}
  c_l = \frac{1}{|S_l|} \sum_{(x^i_l, l, t^i_l) \in S_l} Enc(x^i_l)
\end{equation}
\begin{equation}
  c_a = \frac{1}{|S_a|} \sum_{(x^i_{lw}, l, t^i_{lw}) \in S_a} Enc(x^i_{lw}|x^i_l)
\end{equation}
Here $c_l$ and $c_a$ are the prototype for intent $l$ and slot $a$; $Enc(x^i_l)$ and $Enc(x^i_{lw}|x^i_l)$ are the embeddings of utterance $x^i_l$ and token $x^i_{lw}$ in utterance $x^i_l$ respectively; $w$ is the token index;
\begin{equation}
 S_a = \{(x^i_{lw}, l, t^i_{lw}) | i \in (1 ... k_s), l \in L, w \in (1 ... |x^i_w|), t^i_{lw} = a\}
\end{equation}
is the token level example set for slot $a$. Given an example $(x^{\ast}, l^{\ast}, t^{\ast}) \in Q$, we predict the intent and each token's slot by computing the softmax of distance from the example to prototypes. Specifically,
\begin{equation}
  p(l^{\ast}=l|x^{\ast}, S) = \frac{exp(-|Enc(x^{\ast}) - c_l|^2)}{\sum_{l’} exp(-|Enc(x^{\ast}) - c_{l’}|^2)}
\end{equation}
\begin{equation}
  p(t_w^{\ast}=a|x^{\ast}, S) = \frac{exp(-|Enc(t_w^{\ast}|x^{\ast}) - c_a|^2)}{\sum_{a’} exp(-|Enc(t_w^{\ast}|x^{\ast})-c_{a’}|^2)}
\end{equation}
We denote the approach of building IC and SL classifiers with this framework as \textbf{Proto}.

In the implementation, we jointly pre-train IC and SL. Thus the loss function is defined as the sum of IC and SL negative log-likelihood averaged over instances in $Q^{pretrain}$ given prototypes computed from $S^{pretrain}$. We backward propagate the gradient of loss at pre-training episodically to tune the encoder. At the testing, the learned encoder along with $S^{test}$ is used for calculating class prototype and predict examples in $Q^{test}$.

For comparison, we also build two baselines, one based on MAML (denoted as \textbf{MAML}) and another based on fine-tuning (denoted as \textbf{Finetune}). MAML is another popular meta-learning framework. We utilize MAML to optimize parameters $\theta$ of the IC-SL classifier $f(x | \theta)$ on $S^{pretrain}$ and $Q^{pretrain}$, such that after $\theta$ is adapted with the $S^{test}$ for $d$ steps (i.e., backward update gradient by $d$ epochs), the resulting classifier $f(x | \theta')$ can generalize well in $Q^{test}$. Concretely, we perform the following two-step optimization at pre-training:
\begin{equation}
  \begin{aligned}
    1.& ~~\theta \gets finetune(\theta, d| S)\\
    2.& ~~\theta \gets \theta - \nabla_{\theta}loss(f(\cdot| \theta’), Q)
  \end{aligned}
\end{equation}
Then the learned classifier $f(\cdot| \theta)$ is adapted and evaluated with $S^{test}$ and $Q^{test}$ respectively. Besides, for computational efficiency, in implementation, we adopt first-order approximation of MAML, foMAML, which has been shown to achieve similar performance with less computation \cite{MAML, Jason_Yi_ACL}. On the other hand, \textbf{Finetune} is a common supervised-learning framework for low resource SLU \cite{goyal2018fast}. In \textbf{Finetune}, we pre-train models with examples from {$S^{pretrain}$, $Q^{pretrain}$} in batch. Adaptation and evaluation are also conducted in episodes, where the output layers of pre-trained models are first adapted with $S^{test}$, and then the resulting IC-SL models are evaluated with $Q^{test}$.

\subsection{Model architecture}

Figure \ref{fig:diagram} visualizes the architectures for the IC-SL joint classifiers built with these frameworks. Three popular architectures for SLU, \textbf{ELMo}, \textbf{GloVe}, and \textbf{BERT}, are investigated \cite{Jason_Yi_ACL}. The architectures share the same design of using a bi-LSTM \cite{LSTM} layer to encode embeddings and then fully connected IC and SL prediction layers with bi-LSTM hidden states as input. These classifiers differ in the utilized embeddings, where \textbf{ELMo} adopts a concatenation of GloVe \cite{GloVe} and ELMo \cite{ELMo}, \textbf{GloVe} uses GloVe only and \textbf{BERT} employs a pre-trained BERT for encoding tokens. In our experiments, we keep these pre-trained encoders frozen in adaptation since our previous study shows that it is insufficient to adapt these encoders with few-shot examples \cite{Jason_Yi_ACL}. When applying these architectures to learning framework \textbf{Proto}, the distance between the output and prototypes is further computed, and the probability of an IC or SL class is the softmax of the distance. As for \textbf{MAML} and \textbf{Finetune}, the output of IC and SL layers is used directly as the prediction logits. Note that our learning frameworks discussed above are architecture agnostic. For the generalizability of our experiment, we choose three popular architectures in SLU to explore. Other model backbones are also viable for these frameworks.

\section{Experiments}

\subsection{Datasets and experiment setup}

We build the few-shot learning task to evaluate the proposed approach based on three public SLU datasets: ATIS \cite{ATIS}, SNIPS \cite{SNIPS}, and TOP \cite{TOP}. ATIS is a dataset in the airline domain, while SNIPS comprises utterances in inquiring music, media, and weather. TOP pertains to navigation and event search with nested and flat intent labels. As discussed in previous work \cite{Jason_Yi_ACL}, we only utilize the non-hierarchical intents in experiments for comparable results. In the context of few-shot learning, data for pre-training and adaptation is often in mismatched domains. Hence, we build the few-shot learning datasets, SNIPS-fs, ATIS-fs, and TOP-fs, by manually selecting intents from SNIPS, ATIS, and TOP respectively for test split, and intents from the remaining two datasets for pre-training and validation splits. In Table \ref{tab:tasks}, we provide detailed statistics of these few-shot datasets\footnote{The splits of intents are selected by maximizing the distance between intents belonging to different splits, where each intent is represented by the average over the BERT-CLS embeddings of its utterances. We also investigated different split methods, such as random sampling, and observed no significant difference. We will release the splits and resulting datasets for reproducibility.}.

\begin{table}[]
  \centering
  \setlength\tabcolsep{2pt}
  \renewcommand{\arraystretch}{1.3}
  \fontsize{9.5}{9.5}\selectfont
  \begin{tabular}{c|c|c|c}
  \hline
    Task\textbackslash Splits & Pre-train          & Validation          & Test           \\ \hline
    \multirow{2}{*}{SNIPS-fs} & (20345, 7, TOP)  & (4333, 5, TOP) & (6254, 3, SNIPS) \\
                              & (4373, 5, ATIS)  & (662, 7, ATIS) &                \\ \hline
    \multirow{2}{*}{ATIS-fs}  & (20345, 7, TOP)  & (4333, 5, TOP) & (829, 7, ATIS)   \\
                              & (8230, 5, SNIPS) &              &                \\ \hline
    \multirow{2}{*}{TOP-fs}   & (4373, 5, ATIS)  & (662, 7, ATIS) & (4426, 6, TOP)   \\
                              & (8230, 5, SNIPS) &              &               \\ \hline
  \end{tabular}
  \vspace*{2mm}
  \caption{Statistics of pre-train, validation, and test splits for established few-shot datasets, shown in the form of (utterance counts, intent counts, datasets from which the intents were selected).}
  \label{tab:tasks}
\end{table}

With the established splits, we extract episodes. At the pre-training stage, the number of intents in each episode, i.e., the $n$-way, is sampled from [3, $|L^{pretrain}|$] uniformly. After that, we sample $n$ intents from $L^{pretrain}$, and sample $k_s$ and $k_q$ utterances respectively for each sampled intent as the support and query set. At the validation and test stage, $n$ is set to $|L^{validation}|$ and $|L^{test}|$, and the remaining settings are the same. Additionally, with the same rationale as many meta-learning studies where $k_s$ and $k_q$ are set to some small numbers arbitrarily \cite{MAML, Proto}, in the experiment here, we let both $k_s$ and $k_q$ equal to 10.

Additional steps are required to introduce noises to these few-shot datasets. For adaptation example missing/replacing, we keep the pre-training and validation episodes the same, while perturbing the test episodes by sampling $c$ (set to arbitrarily small numbers, 1 to 5, in the experiment) utterances from each intent in the support set, and \textit{remove} the sampled ones or \textit{replace} with others. For modality mismatch, since SNIPS and TOP only contain human text input, we use commercial TTS\footnote{Amazon Polly: https://aws.amazon.com/polly/} and ASR services\footnote{Amazon Transcribe: https://aws.amazon.com/transcribe/}, to synthesize the audio and decode the audio back to text. Slot labels on the human text are projected onto ASR hypotheses with Levenshtein alignment. We adopt a similar process for ATIS but skip TTS since audio recordings are available in ATIS. The word error rate (WER) of decoded audio for ATIS, SNIPS, and TOP is 18.4\%, 16.2\%, and 14.7\%, respectively.

In the experiment here, we pre-train and adapt all the models using Adam optimizer. The learning rate is set to 0.001 in \textbf{Finetune} and \textbf{Proto}. For \textbf{MAML}, the rate is 0.003 at the pre-training step and 0.01 at the adaptation with $d$ set to 8. We pre-train models for 30 epochs either episodically (\textbf{MAML} or \textbf{Proto}) or in batch (\textbf{Finetune}) with size 512. At the testing, we adapt these models for ten steps on the support set. We select all the hyperparameters for each approach separately, such that the hyperparameters yield the minimum IC and SL joint loss averaged over the three validation sets without perturbation. As for the model architecture, we use a 2-layer BiLSTM with 256 hidden units for contextual encoding. GloVe with 300 dimensions, ELMo with 512 dimensions, and BERT-medium with 512 dimensions are selected for the token embeddings as these settings are commonly adopted in SLU experiments. To assess the model performance, we report the average IC accuracy and SL F1 over 100 episodes and three random seeds, a typical setting in few-shot learning to avoid performance fluctuation \cite{Jason_Yi_ACL}.

\subsection{Experiment results}

Then we evaluate the performance and robustness of the proposed method (\textbf{Proto}) as well as baselines (\textbf{MAML} and \textbf{Finetune}) on established few-shot noisy tasks. Table \ref{tab:robust_1} shows the IC accuracy and SL F1 for the methods under three settings of adaptation example perturbation (i.e., no perturbation, \textit{removing} or \textit{replacing} $c$ utterances). Here, we start from minimal perturbation, $c$ = 1. For evaluating robustness, we also report in parentheses the absolute accuracy and F1 difference between each perturbed setting and no perturbation counterpart\footnote{Note that for measuring the absolute difference, we first calculate the difference in IC/SL performance between the noisy and no perturbation setting for each episode. The average and standard deviation of the difference over episodes are reported in the table. The reported metrics are different from those obtained by directly computing the difference between the averaged performance reported in the table for the noisy and no perturbation settings.}. Similar to previous observation \cite{Jason_Yi_ACL}, we find \textbf{Proto} outperforms \textbf{MAML} and \textbf{Finetune} consistently for IC and SL problems in all the few-shot datasets; also the relatively low SL F1 from baselines suggests that few-shot SL is challenging. In addition to the three learning frameworks, the impact of various model architectures is investigated as well. We observe that \textbf{BERT} consistently yields better performance than \textbf{GloVe} and \textbf{ELMo} (c.f., row 1 to 9), presumably because BERT encodes more knowledge for SLU. Thus, in the following, we choose \textbf{BERT} as the backbone model architecture.

Results in Table \ref{tab:robust_1} also suggest that \textbf{Proto} yields the most robust performance against adaptation example missing/replacing, with absolute differences of IC accuracy and SL F1 between 0.2 and 1.7. We surmise the robustness results from the model's ensemble nature, where the inference can be viewed as an aggregation of classifier prediction based on the distance to examples. \textbf{Finetune} is comparably robust for IC in \textit{replacement} but worse in \textit{removal}, presumably because the latter leaves fewer adaptation utterances, which can be consequential at few-shot learning. The SL performance from \textbf{Finetune} is too low to measure the robustness. \textbf{MAML}, on the other hand, exhibits a large variation in performance. We believe the reason is that the adaptation in MAML, which decides where to evaluate the gradient, amplifies perturbation.

\begin{table*}[t]
  \centering
  \begin{adjustwidth}{-0.2in}{0in}
    \setlength\tabcolsep{3.0pt}
    \begin{tabular}{cccccccc}
        \toprule
        \multicolumn{1}{c}{} & \multicolumn{1}{c|}{} & \multicolumn{3}{c|}{IC acc. (absolute acc. difference)}           & \multicolumn{3}{c}{SL F1 (absolute F1 difference)} \\ \cline{3-8} 
        \multicolumn{1}{c}{} & \multicolumn{1}{c|}{}                          & SNIPS-fs     & ATIS-fs       & \multicolumn{1}{c|}{TOP-fs}        & SNIPS-fs        & ATIS-fs         & TOP-fs          \\ \hline
        \multicolumn{1}{c|}{} & \multicolumn{1}{c|}{\textbf{FINETUNE}~}                  & ~83.6$\pm$0.8  & 69.9$\pm$1.6   & \multicolumn{1}{c|}{57.7$\pm$0.7}   & ~19.6$\pm$0.7      & 20.1$\pm$0.6      &15.7$\pm$0.6      \\
        \multicolumn{1}{c|}{} & \multicolumn{1}{c|}{\textbf{FINETUNE-GloVe}~}                  & ~81.0$\pm$0.7  & 65.6$\pm$1.1   & \multicolumn{1}{c|}{53.1$\pm$0.6}   & ~18.0$\pm$0.4      & 18.2$\pm$0.5      &14.7$\pm$0.4      \\
        \multicolumn{1}{c|}{} & \multicolumn{1}{c|}{\textbf{FINETUNE-ELMo}~}                  & ~82.1$\pm$1.2  & 68.7$\pm$1.9   & \multicolumn{1}{c|}{58.0$\pm$1.0}   & ~19.3$\pm$0.6      & 19.4$\pm$0.8      &14.9$\pm$0.4      \\
        \multicolumn{1}{c|}{No} & \multicolumn{1}{c|}{\textbf{MAML}}       & ~87.4$\pm$0.8  & 71.1$\pm$1.1   & \multicolumn{1}{c|}{57.6$\pm$0.5}   & ~33.0$\pm$1.3     & 26.3$\pm$0.9      & 22.9$\pm$0.8      \\
    
        \multicolumn{1}{c|}{perturbation} & \multicolumn{1}{c|}{\textbf{MAML-GloVe}}                      & ~79.4$\pm$0.9  & 65.9$\pm$0.9   & \multicolumn{1}{c|}{53.2$\pm$0.6}   & ~30.1$\pm$1.0     & 24.9$\pm$0.8      & 21.3$\pm$0.6      \\
        \multicolumn{1}{c|}{} & \multicolumn{1}{c|}{\textbf{MAML-ELMo}}                      & ~83.5$\pm$0.9  & 69.6$\pm$1.3   & \multicolumn{1}{c|}{54.5$\pm$0.8}   & ~31.9$\pm$1.1     & 25.4$\pm$1.2      & 23.2$\pm$0.9      \\
        \multicolumn{1}{c|}{} & \multicolumn{1}{c|}{\textbf{Proto}}                     & ~\textbf{90.9$\pm$0.3}  & 75.3$\pm$0.7   & \multicolumn{1}{c|}{\textbf{61.9$\pm$1.1}}   & ~\textbf{45.4$\pm$0.5}     & \textbf{42.7$\pm$1.6}     & 35.4$\pm$0.5     \\
        \multicolumn{1}{c|}{} & \multicolumn{1}{c|}{\textbf{Proto-GloVe}}    & ~75.3$\pm$0.4  & 70.1$\pm$0.6   & \multicolumn{1}{c|}{52.7$\pm$1.3}   & ~32.2$\pm$0.7     & 41.5$\pm$1.5     & 30.6$\pm$0.6     \\
        \multicolumn{1}{c|}{} & \multicolumn{1}{c|}{\textbf{Proto-ELMo}}     & ~87.1$\pm$0.5  & \textbf{76.0$\pm$0.8}   & \multicolumn{1}{c|}{59.8$\pm$1.2}   & ~43.1$\pm$0.6     & 41.2$\pm$1.8     & \textbf{35.7$\pm$0.8}     \\ \hline
        \multicolumn{1}{c|}{} & \multicolumn{1}{c|}{\multirow{2}{*}{\textbf{FINETUNE}~}} & ~83.4$\pm$0.6  & 69.3$\pm$1.4   & \multicolumn{1}{c|}{57.4$\pm$0.6}   & ~18.9$\pm$0.7      & 19.8$\pm$0.7      & 15.3$\pm$0.4      \\
        \multicolumn{1}{c|}{\textit{Remove} 1} & \multicolumn{1}{c|}{}                          & ~(3.1$\pm$3.0) & (4.1$\pm$3.3)  & \multicolumn{1}{c|}{(4.0$\pm$2.9)}  & ~\textbf{(1.1$\pm$1.8)}    & (1.1$\pm$1.4)    & (0.9$\pm$0.6)    \\
        \multicolumn{1}{c|}{utterance} & \multicolumn{1}{c|}{\multirow{2}{*}{\textbf{MAML}}}     & ~87.3$\pm$0.5  & 71.3$\pm$1.6   & \multicolumn{1}{c|}{58.8$\pm$1.4}   & ~32.6$\pm$1.4     & 25.9$\pm$0.8      & 22.3$\pm$0.7      \\
        \multicolumn{1}{c|}{per intent} & \multicolumn{1}{c|}{}                          & ~(4.3$\pm$4.2) & (11.5$\pm$8.3) & \multicolumn{1}{c|}{(10.7$\pm$8.8)} & ~(5.2$\pm$4.0)    & (3.8$\pm$2.8)    & (4.4$\pm$3.5)    \\
        \multicolumn{1}{c|}{} & \multicolumn{1}{c|}{\multirow{2}{*}{\textbf{Proto}}}    & ~90.7$\pm$0.3  & 75.0$\pm$0.6   & \multicolumn{1}{c|}{61.3$\pm$0.9}   & ~44.8$\pm$0.6     & 42.5$\pm$1.5     & 35.2$\pm$0.5     \\
        \multicolumn{1}{c|}{} & \multicolumn{1}{c|}{}                          & ~\textbf{(0.8$\pm$1.6)} & \textbf{(0.3$\pm$0.6)}  & \multicolumn{1}{c|}{\textbf{(1.3$\pm$1.4)}}  & ~(1.5$\pm$1.5)    & \textbf{(0.4$\pm$0.5)}    & \textbf{(0.2$\pm$0.3)}    \\ \hline
        \multicolumn{1}{c|}{} & \multicolumn{1}{c|}{\multirow{2}{*}{\textbf{FINETUNE}~}} & ~83.2$\pm$0.9  & 69.4$\pm$1.5   & \multicolumn{1}{c|}{57.1$\pm$0.8}   & ~18.8$\pm$0.5      & 20.0$\pm$0.6      & 15.5$\pm$0.5      \\
        \multicolumn{1}{c|}{\textit{Replace} 1} & \multicolumn{1}{c|}{}                          & ~(1.9$\pm$2.4) & (0.6$\pm$1.0)  & \multicolumn{1}{c|}{\textbf{(1.6$\pm$1.5)}}  & ~\textbf{(1.7$\pm$1.5)}    & (1.1$\pm$0.5)    & (1.0$\pm$0.2)    \\
        \multicolumn{1}{c|}{utterance} & \multicolumn{1}{c|}{\multirow{2}{*}{\textbf{MAML}}}     & ~87.5$\pm$0.7  & 71.0$\pm$1.7   & \multicolumn{1}{c|}{57.5$\pm$1.3}   & ~32.8$\pm$1.3     & 25.6$\pm$0.1      & 22.8$\pm$0.6      \\
        \multicolumn{1}{c|}{per intent} & \multicolumn{1}{c|}{}                          & ~(2.0$\pm$2.7) & (5.3$\pm$5.0)  & \multicolumn{1}{c|}{(4.9$\pm$5.4)}  & ~(3.2$\pm$2.6)    & (1.8$\pm$1.7)    & (1.5$\pm$2.0)    \\
        \multicolumn{1}{c|}{} & \multicolumn{1}{c|}{\multirow{2}{*}{\textbf{Proto}}}    & ~90.9$\pm$0.4  & 75.2$\pm$0.6   & \multicolumn{1}{c|}{62.1$\pm$1.0}   & ~45.5$\pm$0.6     & 42.6$\pm$1.5     & 35.1$\pm$0.5     \\
        \multicolumn{1}{c|}{} & \multicolumn{1}{c|}{}                          & ~\textbf{(0.9$\pm$1.8)} & \textbf{(0.4$\pm$0.7)}  & \multicolumn{1}{c|}{(1.7$\pm$1.7)}  & ~\textbf{(1.7$\pm$1.6)}    & \textbf{(0.6$\pm$0.6)}    & \textbf{(0.3$\pm$0.4)}    \\ \bottomrule
    \end{tabular}
  \end{adjustwidth}
  \caption{Average IC accuracy and SL F1 over 100 test episodes for three few-shot datasets in the form of mean $\pm$ standard deviation, computed over three random seeds. We show results for three methods and perturbation settings. In the parentheses, the absolute accuracy and F1 difference on test episodes between \textit{replace} or \textit{remove} and no perturbation counterpart is reported. Model architecture \textbf{BERT} is used by default if not specified.}
  \label{tab:robust_1}
\end{table*}

\begin{figure}[t]
  \centering
  \includegraphics[width=.7\linewidth]{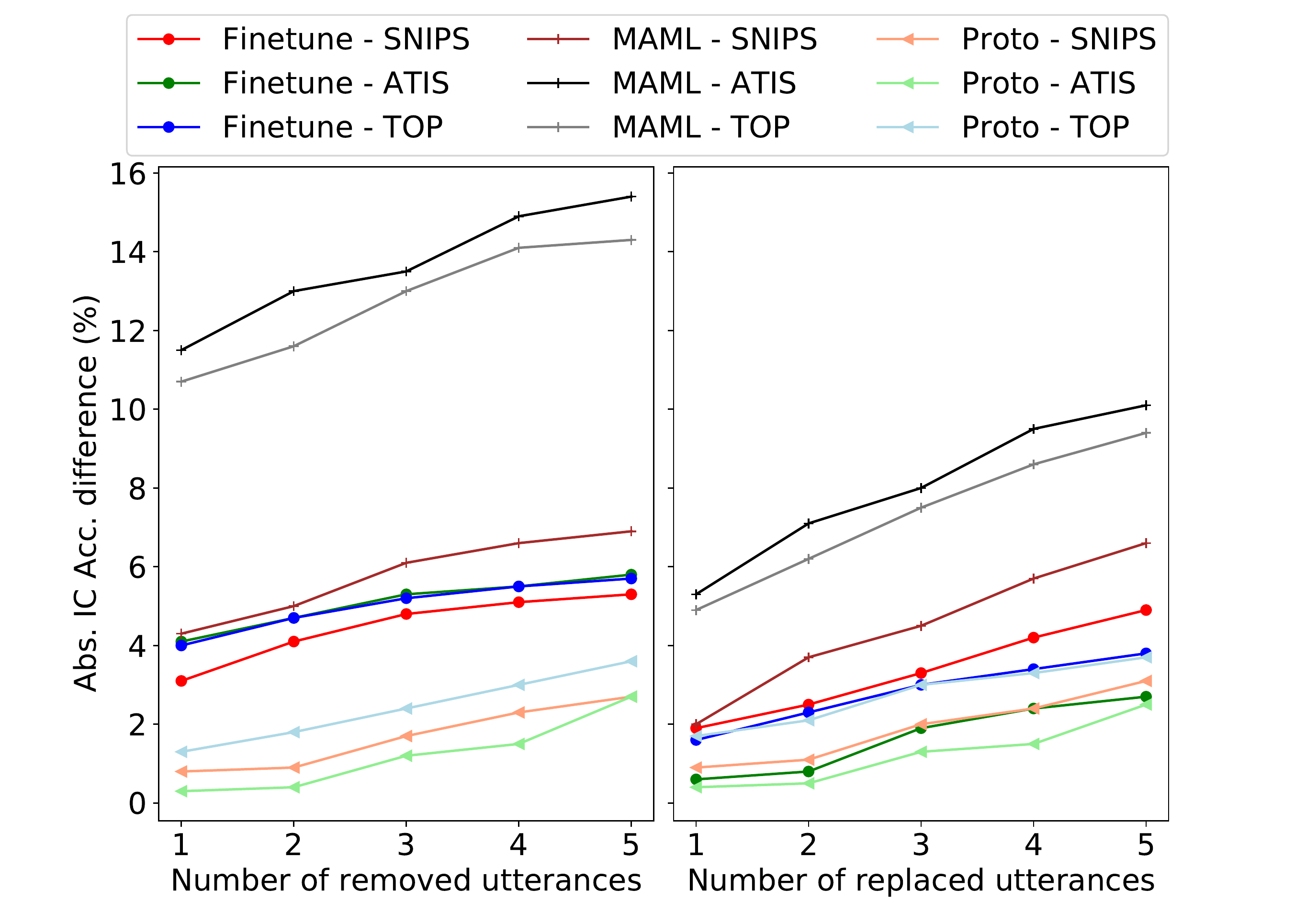}
  \caption{The absolute IC accuracy difference between no perturbation and perturbed settings with various $c$, the number of utterances \textit{removed} or \textit{replaced} from the adaptation set.}
  \label{fig:robust_plot}
\end{figure}

We further vary $c$, the number of perturbed examples, and quantify its performance impact. In Figure \ref{fig:robust_plot}, we present the absolute difference in IC accuracy between perturbed and no perturbation setting over various $c$ (we decided not to look into SL robustness since only \textbf{Proto} yields satisfactory results even before perturbation). \textit{Remove} and \textit{replace} perturbation is shown in the left and right panel respectively. Again \textbf{Proto} yields the most robust results (in terms of least difference) compared to baselines in explored perturbation settings. Also, utterance \textit{removing} is more challenging for model robustness. Both findings strengthen our argument above.

Lastly, we measure the model robustness against modality mismatch. In Table \ref{tab:robust_2}, we report the IC accuracy and SL F1 when models are pre-trained and adapted in human transcription while evaluated with ASR hypotheses. Similar to what we found above, \textbf{Proto} yields the best performance in both IC and SL. By comparing the results in the condition of mismatched modality reported here with the matched modality counterpart (i.e., no perturbation in Table \ref{tab:robust_1}), we observe that \textbf{Proto} is again the most robust approach in IC (accuracy drop ranging from 0.3 to 2.0 for \textbf{Proto}, 3.0 to 4.4 for \textbf{Finetune}, and 3.5 to 4.3 for \textbf{MAML}). SL result from \textbf{Proto} is stable as well (2.2 to 4.3 F1 drop), while \textbf{Finetune} and \textbf{MAML} yield relatively low F1 scores. Findings here agree with the observation made above for adaptation example missing/replacing, and further support our discussion about the robustness of different learning frameworks.

\begin{table}[]
  \centering
  \begin{tabular}{ccccc}
    \toprule
    \multicolumn{1}{c}{} & \multicolumn{1}{c|}{} & SNIPS-fs    & ATIS-fs     & TOP-fs  \\ \hline
    \multicolumn{1}{c|}{} & \multicolumn{1}{c|}{\textbf{FINETUNE}} & 79.4$\pm$1.8 & 66.9$\pm$1.9 & 53.3$\pm$1.1  \\
    \multicolumn{1}{c|}{IC acc.} & \multicolumn{1}{c|}{\textbf{MAML}}     & 83.3$\pm$0.9 & 68.1$\pm$0.9 & 54.0$\pm$1.0 \\
    \multicolumn{1}{c|}{} & \multicolumn{1}{c|}{\textbf{Proto}}    & \textbf{89.0$\pm$0.7} & \textbf{75.0$\pm$0.7} & \textbf{59.9$\pm$0.6} \\ \hline
    \multicolumn{1}{c|}{} & \multicolumn{1}{c|}{\textbf{FINETUNE}}  & 17.5$\pm$0.4  & 19.2$\pm$0.2  & 14.5$\pm$0.3  \\
    \multicolumn{1}{c|}{SL F1} & \multicolumn{1}{c|}{\textbf{MAML}}      & 26.7$\pm$1.6 & 20.7$\pm$0.5  & 18.8$\pm$0.5  \\
    \multicolumn{1}{c|}{} & \multicolumn{1}{c|}{\textbf{Proto}}     & \textbf{41.1$\pm$0.6} & \textbf{39.7$\pm$1.3} & \textbf{33.2$\pm$0.6} \\
    \bottomrule
  \end{tabular}
  \vspace*{2mm}
  \caption{IC accuracy and SL F1 from models pre-trained and adapted in human transcription while evaluated with ASR hypotheses.}
  \label{tab:robust_2}
\end{table}

\section{Conclusions}

In this paper, we establish a novel SLU task, the few-shot noisy SLU, with existing public datasets. We further propose a ProtoNets based approach, \textbf{Proto}, to build IC and SL classifiers with few noisy examples. When there is no noise in few-shot examples, \textbf{Proto} yields better performance than other approaches utilizing MAML and fine-tuning frameworks. \textbf{Proto} also achieves the highest and most robust IC accuracy and SL F1 when two types of noise, adaptation example missing/replacing and modality mismatch, are injected in adaption and evaluation set respectively. We believe the ensemble nature of ProtoNets benefits the model robustness, and the simplicity of \textbf{Proto}'s model architecture is also helpful in the few-shot noisy scenario. Our contribution here is a step toward the efficient and robust deployment of SLU models. While our results are promising, there is still substantial work, from the creation of few-shot SLU datasets covering more noises to studies of faster and stabler learning algorithms, in pursuit of the goal.

\bibliography{main}

\bibliographystyle{abbrvnat}

\newpage

\appendix
\renewcommand{\thesection}{\Alph{section}}
\end{document}